\documentclass[letterpaper]{article} 
\usepackage[]{aaai25}  
\usepackage{times}  
\usepackage{helvet}  
\usepackage{courier}  
\usepackage[hyphens]{url}  
\usepackage{graphicx} 
\usepackage{natbib}  
\usepackage{multirow}
\usepackage{tikz}
\usepackage{caption} 
\usepackage{amsmath}
\usepackage{algorithm}
\usepackage{algpseudocode}
\usepackage{booktabs}
\usepackage{amssymb}
\newcommand{\minisection}[1]{\vspace{0.05in}\noindent {\bf #1}}

\usepackage{newfloat}
\usepackage{listings}
\DeclareCaptionStyle{ruled}{labelfont=normalfont,labelsep=colon,strut=off} 
\floatstyle{ruled}
\newfloat{listing}{tb}{lst}{}
\floatname{listing}{Listing}
\title{MeRino: Entropy-Driven Design for Generative Language \\ Models on IoT Devices}

\title{MeRino: Entropy-Driven Design for Generative Language Models on IoT Devices}
\author {
    Youpeng Zhao\textsuperscript{\rm 1},
    Ming Lin\textsuperscript{\rm 2},
    Huadong Tang\textsuperscript{\rm 3},
    Qiang Wu\textsuperscript{\rm 3},
    Jun Wang\textsuperscript{\rm 1},    
}
\affiliations {
    \textsuperscript{\rm 1}University of Central Florida,
    \textsuperscript{\rm 2}Independent Researcher,
    \textsuperscript{\rm 3}University of Technology Sydney\\
    youpeng.zhao@ucf.edu, linming04@gmail.com, huadong.tang@student.uts.edu.au, \\
    qiang.wu@uts.edu.au,
    jun.wang@ucf.edu
}

\usepackage{bibentry}

\begin{document}

\maketitle

\begin{abstract}
Generative Large Language Models (LLMs) stand as a revolutionary advancement in the modern era of artificial intelligence (AI). However, scaling down LLMs for resource-constrained hardware, such as Internet-of-Things (IoT) devices requires non-trivial efforts and domain knowledge.
In this paper, we propose a novel information-entropy framework for designing mobile-friendly generative language models. 
The whole design procedure involves solving a mathematical programming (MP) problem, which can be done on the CPU within minutes, making it nearly zero-cost. 
We evaluate our designed models, termed MeRino, across fourteen NLP downstream tasks, showing their competitive performance against the state-of-the-art autoregressive transformer models under the mobile setting. Notably, MeRino achieves similar or better performance on both language modeling and zero-shot learning tasks, compared to the 350M parameter OPT while being 4.9$\times$ faster on NVIDIA Jetson Nano with 5.5$\times$ reduction in model size.
\end{abstract}

%

\section{Introduction}
The Transformer architecture, originally introduced in~\citep{attention}, has revolutionized the field of natural language processing (NLP). It has become the de-facto building block in many pre-trained generative large language models (LLMs)~\cite{gpt2,gpt3}. 
Thanks to their ability to scale up to billion-level parameters, LLMs have exhibited exceptional abilities in solving complex tasks through text generation, with prominent applications such as ChatGPT~\cite{chatgpt} and Claude~\cite{cluade}. 
Nevertheless, running LLMs in cloud computing platforms with specialized hardware accelerators can be very expensive.
It is estimated that running a single ChatGPT query might consume around 0.3 kWh, which is roughly 1000$\times$ more than a simple Google search, highlighting the significant financial and environmental impact of recurring LLM usage~\cite{query-energy}.
Therefore, it is imperative to downsize the footprints of LLMs to meet the sustainability and efficiency requirements of ethical and responsible AI solutions~\cite{sustainableai}.

With the ever-increasing popularity of edge computing, deploying LLMs on resource-constrained hardware, e.g., mobile phones and internet-of-things (IoT) devices becomes a much more appealing solution.
Such on-device AI both speeds up response latency and enhances data privacy and security, making LLMs more accessible, efficient, and practical in a wide range of daily applications~\cite{appleai,copilot}. 

\begin{figure}[!t]
     \centering
         \includegraphics[width=\linewidth]{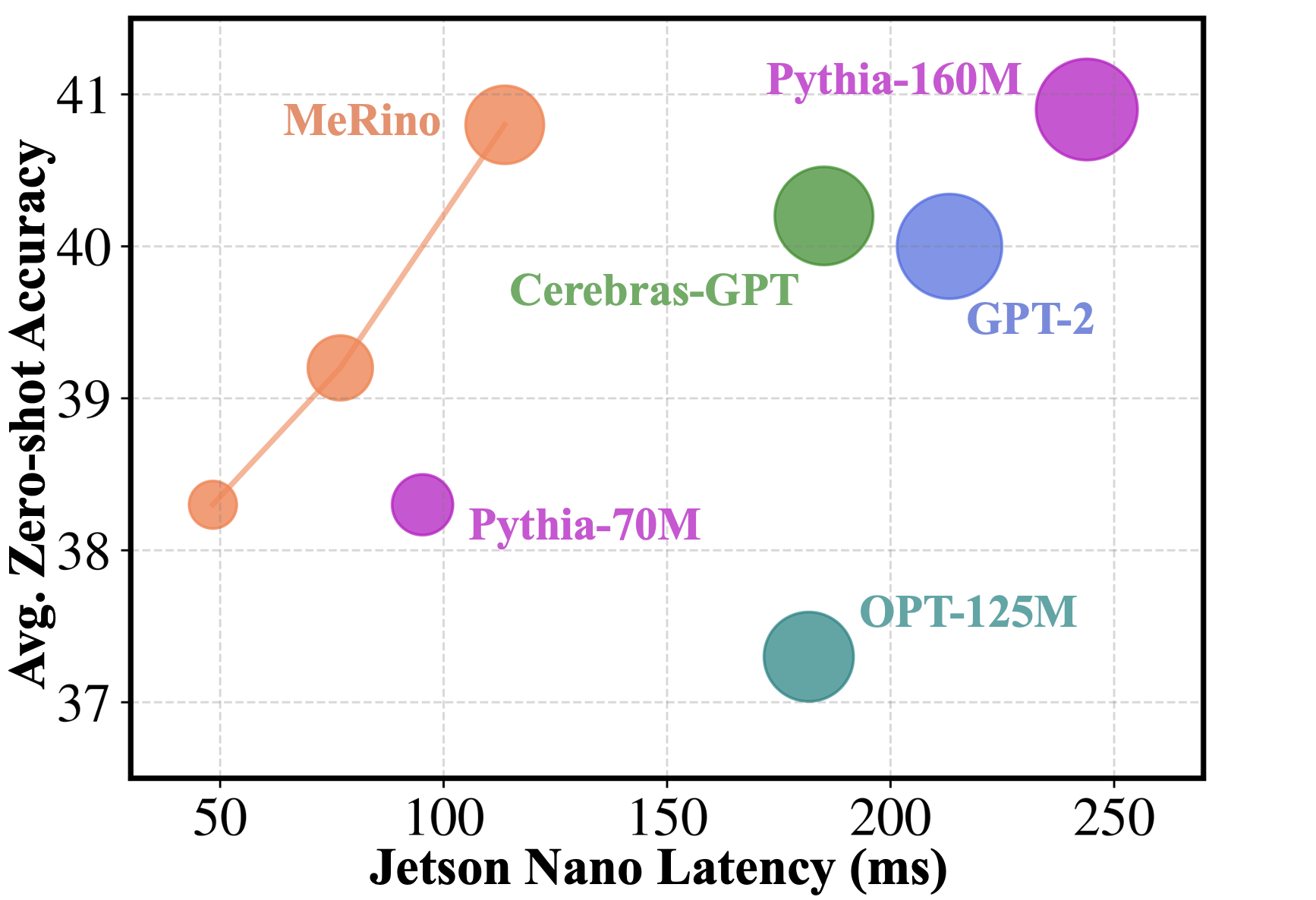}
         \vspace{-7mm}
        \caption{Average zero-shot accuracy and inference latency on NVIDIA Jetson Nano for mobile-level LLMs.
        Results were evaluated using lm-evaluation-harness~\cite{lm-eval} on open-sourced pre-trained models.      
        The diameter of each circle denotes the corresponding model FLOPs.
        }
     \label{fig:intro}
\end{figure}
However, the practical deployment of LLMs for edge devices has met with its unique obstacles. 
\underline{First}, integrating existing pre-trained LLMs can be prohibitively expensive, as the weights for most models can easily exceed the main memory capacity. 
For instance, for the OPT-125M model, running a single query of length 128 would consume over 1,200 MB of DRAM or flash memory, with close to 200 ms latency on NVIDIA Jetson Nano. 
For mobile applications, the ideal memory usage should be less than 10\% of the total capacity, to avoid potential system overloading and out-of-memory (OOM) failure.
Several works~\cite{MobileLLM,minicpm} have considered designing optimized LLMs in the regime of 1B parameter, which yields better task performances but suffers from the same latency overhead as previous methods.
This motivates us to deploy even smaller LLMs ($<$100\,M) with even lower latency ($<$100\,ms).
\underline{Second}, the hardware configuration for each edge computing platform varies from device to device. 
Thus, designing a one-fits-all model that satisfies all requirements is a non-trivial task.
Neural architecture search (NAS)~\cite{nasbert}, has recently emerged as a promising solution for efficient architecture designs that offer better accuracy and computation trade-offs against manually designed models.
However, NAS methods are usually computationally expensive, due to the costly model evaluations and supernet-training processes, which often consume thousands of GPU hours. 
To mitigate the high costs and promote the search efficiency of NAS, a series of proxies~\cite{lts,transformerfree} have been proposed to evaluate the accuracy of neural network architectures in a low-cost manner. 
However, these methods generally focus on vision models, and directly applying them to language models could produce sub-optimal solutions. 
Moreover, these methods usually require forward and backward passes over the architecture, which could be compute-intensive for low-end hardware.

In this work, we present an entropy-driven framework to design lightweight variants of generative language models tailored for resource-constrained devices \textbf{under 100\,M parameters}.
Our key idea is to maximize the entropy for autoregressive transformers, which positively correlates with the model performance, under given computational constraints, such as parameter size, FLOPs, and latency.
The optimal model configuration is generated by solving a mathematical programming (MP) problem utilizing an Evolutionary Algorithm (EA).
Our entropy-driven algorithm can automatically design an optimal transformer model running on the target IoT hardware within minutes, significantly reducing the design costs of existing NAS-based methods.

Our designed models termed \textbf{MeRino}, achieve competitive performance compared to OPT and GPT models, with much better parameter and computation efficiency across numerous NLP downstream tasks. 
Notably, MeRino obtains comparable accuracy performance against OPT-350M, with $5.5 \times$ reduction in model size, $4.5 \times$ reduction in FLOPs, and 4.9$\times$ faster latency on NVIDIA Jetson Nano.
MeRino also significantly outperforms previous NAS-based methods by a clear margin, in terms of both model accuracy and search efficiency.

The key contributions of this work are summarized as follows:
\begin{itemize}
    \item We propose an entropy-driven framework to address the challenge of designing efficient generative language models for resource-constrained devices at nearly zero cost.
    \item Our design paradigm leverages the Maximum Entropy Principle and constrained mathematical programming (MP) to optimize transformer architectures given computation budgets.
    \item Experimental results show that our designed models, termed MeRino, achieve much better accuracy/computation tradeoffs against both manually designed and NAS-designed models, with improved parameter, computation, and latency speedup on mobile devices.
\end{itemize}

\section{Related Work}
\minisection{Large Language Models (LLMs).} 
Generative large language models (LLMs) have emerged as the standard solution to a wide range of NLP tasks. 
GPT-3~\cite{gpt3}, in particular, pushed the boundaries of casual language models by scaling up the model size to 175 billion parameters.
In the pursuit of democratizing and fostering reproducible research in LLMs, a series of open-sourced family models are subsequently released, most notably OPT~\cite{opt}, LLaMA~\cite{llama}, Pythia~\cite{pythia}, and Cerebras-GPT~\cite{cerebras}.
MiniCPM~\cite{minicpm} and MobileLLM~\cite{MobileLLM} further unveil the potential of LLMs under mobile settings through various training and design strategies.
However, \textbf{the above-mentioned models are all above 100\,M parameter size}; in this work, we aim to design transformer-based language models in sub-100\,M regime that achieves competitive performance for IoT devices with limited memory space and compute power.

\minisection{Model Compression.} Two of the most widely studied techniques in designing compressed versions of language models are quantization and sparsification. 
Quantization~\cite{GPTQ} reduces the memory footprint by quantizing the model weights from high precision (FP16/32) to low precision (INT4/8) at the cost of negligible accuracy performance drop. 
Sparsity-driven methods aim to remove unnecessary computation by identifying the most crucial components in either model weights~\cite{sparsegpt,slicegpt} or Key-Value (KV) cache~\cite{h2o,alisa}, to accelerate the LLM inference process.
In this work, orthogonal to the above-mentioned methods, our proposed approach emphasizes the architecture design of lightweight transformer-based LLMs.

\minisection{Neural Architecture Search (NAS).} 
Due to its success in computer vision (CV), neural architecture search (NAS) has recently gained attention in the NLP community. 
The general approach is to train a giant supernet~\cite{nasbert} to efficiently search for compressed language model versions, however, it usually incurs heavy computation costs and requires extensive engineering efforts. 
Training-free proxies have been proposed to reduce the evaluation costs of existing NAS methods. 
TE-NAS~\cite{tenas} calculates the neural tangent kernel (NTK) score to estimate the architecture accuracy and TF-TAS~\cite{transformerfree} proposes a gradient-based DSS-Score to rank the architectures in Vision Transformers (ViTs). 
LTS~\cite{lts} utilizes the decoder parameter count as a proxy for perplexity ranking in generative language models. 
However, these methods are generally data-dependent, requiring storing network parameters and weights in memory, which could be compute-intensive on IoT devices with limited power. 
In this work, we propose an entropy-based framework that can design better model architecture and run directly on target hardware with higher search efficiency.

\minisection{Information Theory in Deep Learning.} 
Information theory recently has emerged as a powerful tool for studying deep neural networks. 
Several previous studies~\cite{redunet,info1} have attempted to establish a connection between the information entropy and the neural network architectures. 
For instance, ~\cite{redunet} tries to interpret the learning ability of deep neural networks using subspace entropy reduction. 
~\cite{info1} investigates the information bottleneck in deep architectures and explores the entropy distribution and information flow in deep neural networks.
Additionally, ~\cite{maedet} focuses on designing high-performance vision models via maximizing multi-level entropy. 
In this work, we focus on using information entropy to design efficient generative transformer language models.

\section{Methodology} 
In this section, we begin by presenting some preliminary details on transformer models. 
Next, we introduce our novel definition of network entropy for transformer models. 
Moreover, we demonstrate that the subspace entropy positively correlates with the model performance after training. 
Finally, we present our entropy-driven design procedure, which solves a constrained mathematical programming problem using the Evolutionary Algorithm (EA).
\subsection{Preliminaries}

\begin{table*}[!t]
\caption{Perplexity comparison of two different structures of autoregressive transformer models on the LM1B dataset.
Lower perplexity is better.}
\label{tab:ratio}
\centering
\vspace{-3mm}
\begin{tabular}{cccccccc}
\toprule
Model & $L$ & $E$ & Params & Entropy & Effective $\gamma$ & Entropy w/ $\gamma$ & Perplexity \\
\midrule
`Wide' & 1 & 256 & 40\,M & 2784 & \textbf{0.008} & \textbf{2243} & \textbf{53.7} \\
`Deep' & 24 & 64 & 40\,M & \textbf{4680} & 0.25 & 2042 & 71.9 \\
\bottomrule
\end{tabular}
\end{table*}

\minisection{Multi-Head Self-Attention (MHSA).} Multi-head attention (MHSA) is a crucial component within the transformer architecture that enables the model to selectively attend to different segments of the input sequence. 
This mechanism involves projecting the input sequence into multiple attention heads, each of which calculates an independent attention distribution. 
In MHSA computation, there are specifically four main matrices involved: attention matrices $W^{Q}, W^{K}, W^{V} \in \mathbb{R}^{d_{in} \times d_{in}/h}$ and the final output project matrix $W^{O}\in \mathbb{R}^{d_{in} \times d_{out}}$. 
Given the output of previous layers $X\in \mathbb{R}^{n \times d_{in}}$ as input, the attention function is formulated as:

\begin{align}
    Q, K, V &= XW^{Q}, XW^{K}, XW^{V} \\
    \text{Attn}(Q, K, V) &= \text{SoftMax}(\frac{Q K^{T}}{\sqrt{d_{in}/h}})(V)
\end{align}
where $Q$, $K$, and $V$ represent queries, keys, and values, respectively. MHSA is defined by concatenating $h$ attention heads and producing outputs as follows:
\begin{align}
    \text{MHSA}(X) &= \text{Concat} (\text{Attn}_{i},...,\text{Attn}_{h}) W^{O}
\end{align}
In addition, the transformer layer adopts residual connection and layer normalization on top of MHSA to compute the final outputs.
\begin{align}
    X^{\text{MHSA}} = \text{LayerNorm}(X + \text{MHSA}(X))
\end{align}

\minisection{Position-wise Feed-forward Network (FFN).}
In addition to the MHSA, each transformer layer includes a feed-forward network (FFN). 
The FFN applies two point-wise fully connected layers followed by a non-linear activation function, such as ReLU. 
Operations within FFN can be formulated as follows:
\begin{align}
    X^{\text{FFN}} = \text{ReLU}( X^{\text{MHSA}} W^{\text{FFN}_1}+b_1)W^{\text{FFN}_2} + b_2
\end{align}
Similarly, the FFN also incorporates residual connections and layer normalization to compute the final outputs:
\begin{align}
X^{\text{FFN}} = \text{LayerNorm}(X^{\text{MHSA}} + X^{\text{FFN}})
\end{align}

\subsection{Entropy of Neural Network}
From the perspective of information theory~\citep{mae}, deep neural networks (DNNs) can be regarded as information systems, and their performance is closely related to the expressive power of such networks. 
The notion of entropy is often used to measure such expressiveness in DNNs. 
The fundamental operation in DNNs is matrix multiplication, which computes the weighted sum of inputs and weights in each layer of a neural network. 
According to previous works~\citep{redunet}, for a simple one-layer neural network with weight matrix $W \in \mathbb{R}^{c_1 \times c_2}$, the entropy can be defined as 

\begin{equation}
    \widehat{H}(W) \triangleq \mathbb{E} \{ \sum_{j=1}^{r_i} \log (1+ \frac{s_j^2}{\epsilon^2}) \} \label{eq:1}
\end{equation}
where $r_i=\min(c_1, c_2)$, $s_j$ is the $j$-th largest singular value of $W$ and $\epsilon$ is a small constant. 

For an $L$-layer network $f(\cdot)$, we can define the network entropy by accumulating the entropy of each layer as:
\begin{align}
\widehat{H}_{f} = \widehat H(W_1, W_2, ..., W_L) = \sum_{i=1}^L \widehat H(W_i) \label{eq:2}
\end{align}
The entropy measures the \textit{expressiveness} of the deep neural network, which is positively correlated with the network performance~\cite{maedet}. However, directly maximizing the above-defined entropy leads to the creation of over-deep networks, since according to Eq.~(\ref{eq:2}), the expressivity (entropy) grows exponentially faster in depth (number of layers $L$), than in width (dimension of $W_i$). For an over-deep network, a small perturbation in low-level layers of the network will lead to an exponentially large perturbation in the high-level output of the network~\cite{dlt}. During the back-propagation process, the gradient flow often cannot effectively propagate through the entire network.

\subsection{Effectiveness of Neural Network}
To verify the negative impact when the network is over-deep, in Table \ref{tab:ratio}, we conduct experiments of training two transformer architectures with a similar parameter size of 40\,M. One model, referred to as the `Wide' model, consists of only one layer and an embedding dimension of 256. The other model, referred to as the `Deep' model, consists of 24 layers but only with an embedding dimension of 64. Both models are trained under the same setting until convergence. We observe that even though the `Deep' network has much higher entropy, it obtains worse perplexity performance after training than the `Wide' network. This observation aligns with the common belief that over-deep networks hinder effective information propagation and are difficult to train and optimize~\cite{dlt}. 

To address the potential trainability issues, we propose adding additional constraints to control the depth-width ratio of networks. Specifically, we adopt the term \textit{effectiveness} $\gamma$ from the work~\citep{dlt} and define it as follows:
\begin{align}
    \gamma = {\beta L} / \hat{w}
    \label{eq:3}
\end{align}
Here, $\hat{w}$ is the effective width of a $L$-layer network and $\beta$ is a scaling factor to control $\gamma$ within the range of 0 and 1. To enforce the above constraint, we revise Eq.~(\ref{eq:3}) as follows:
\begin{align}
    \widehat{H}_{f} = (1-\gamma)\sum_{i=1}^L H(W_i) \label{eq:4}
\end{align}
Compared to the previous subspace entropy definition, Eq.~(\ref{eq:4}) penalizes networks with larger depth-to-width ratios (higher $\gamma$). 
This constraint helps alleviate potential trainability issues by promoting a more balanced depth-width ratio in the network architecture. 
By considering both \textit{expressiveness} (entropy) and \textit{effectiveness} (the depth-width ratio), we aim to design more capable and trainable models.

\subsection{Entropy of Transformers}
Consider a $L$-layer transformer model with embedding dimension $E$ and FFN dimension $F$, according to Theorem 1 in ~\citep{dwi}, the depth-width sufficiency behavior satisfied a logarithmic condition in transformer models. 
Subsequently, we propose to define the effective width of MHSA and FFN and their corresponding entropy as:
\begin{align}
\hat{w}_{\text{MHSA}} &= \log E, \quad \quad \hat{w}_{\text{FFN}} = \log F\\
\widehat{H}_{\text{MHSA}}&= (1-\frac{\beta L}{\hat{w}_{\text{MHSA}}}) \sum_{i=1}^L \widehat{H}(W_i^{Q},W_i^{K},W_i^{V},W_i^{O}) \\
\widehat{H}_{\text{FFN}} &= (1-\frac{\beta L}{\hat{w}_{\text{FFN}}})\sum_{i=1}^L \widehat{H}(W_i^{\text{FFN}_1},W_i^{\text{FFN}_2}) 
\label{eq:6}
\end{align}
Therefore, we define the total entropy of the transformer model as linear combinations of the MHSA and FFN entropy:
\begin{align}
\widehat{H} = \alpha_1 \widehat{H}_{\text{MHSA}} + \alpha_2 \widehat{H}_{\text{FFN}}   
\end{align} 
where ${\alpha} = (\alpha_1, \alpha_2)$ are tunable hyperparameters.

\begin{figure}[!t]
     \centering
         \includegraphics[width=.9\linewidth]{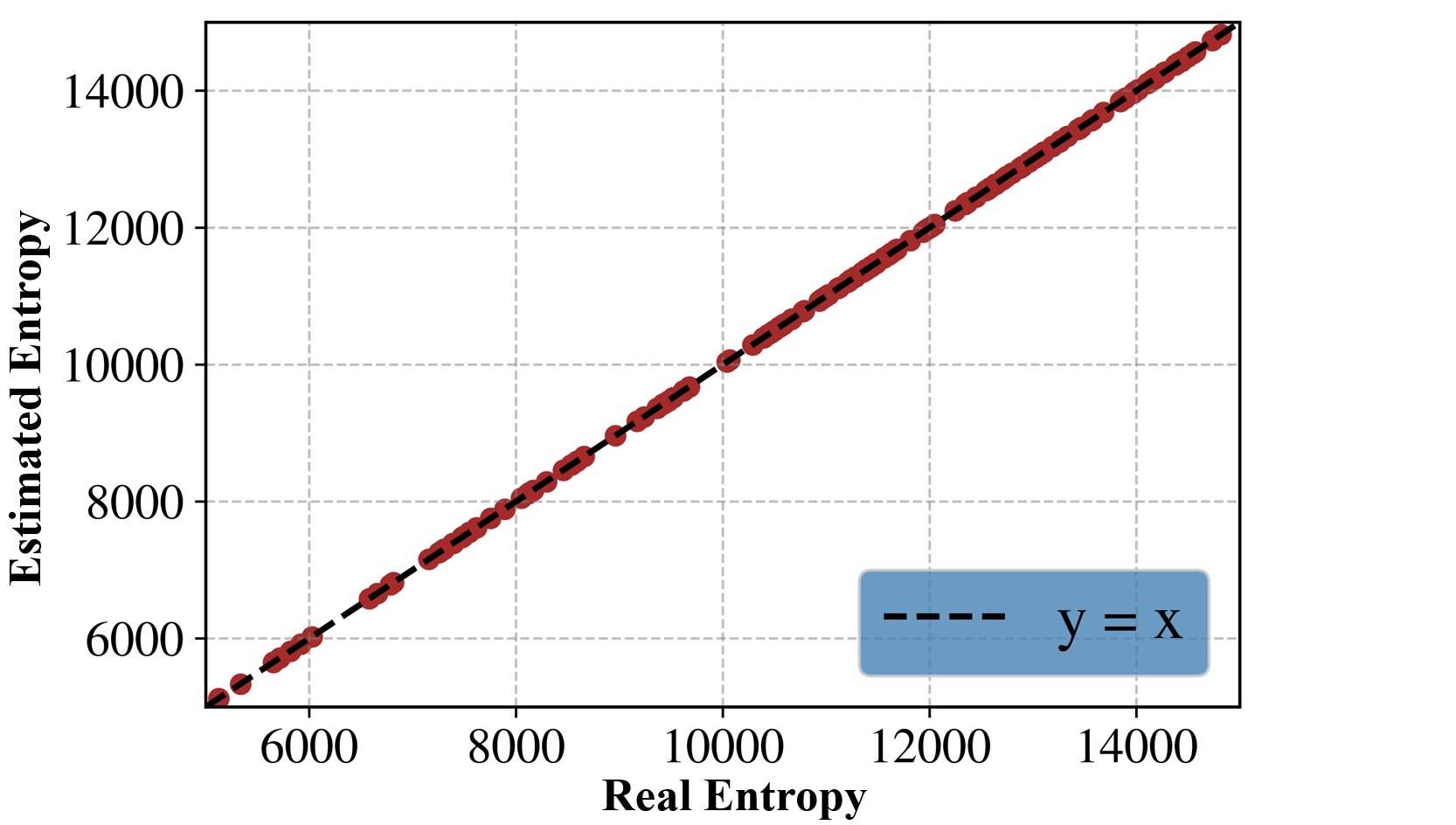}
         \vspace{-3mm}
        \caption{Our entropy estimation based on table lookup is very accurate, with an average error rate of 0.03\%.
        }
     \label{fig:tabel}
\end{figure}

\minisection{Fast Entropy Approximation.} Given the above definitions, we can easily calculate entropy for any transformer model.
However, performing singular value decomposition (SVD) is a costly operation. 
For large models, it sometimes requires minutes to run SVD, which inhibits an efficient design. 
To accelerate the entropy computation, we build an entropy lookup table to approximate the total entropy of a given transformer model. 
The lookup table is built through a pre-computation process that considers all possible combinations of expected entropy values for different dimensions. 
This step incurs only a one-time cost and the resulting lookup table can be shared across multiple experiments.
With the lookup table in place, we can efficiently calculate the entropy of transformer models and enable a more efficient design process for transformer models.
Figure~\ref{fig:tabel} shows that our table lookup method can provide very accurate entropy estimation at almost zero cost, thus greatly speeding up the search process.

\begin{figure*}[!t]
     \centering
         \includegraphics[width=\linewidth]{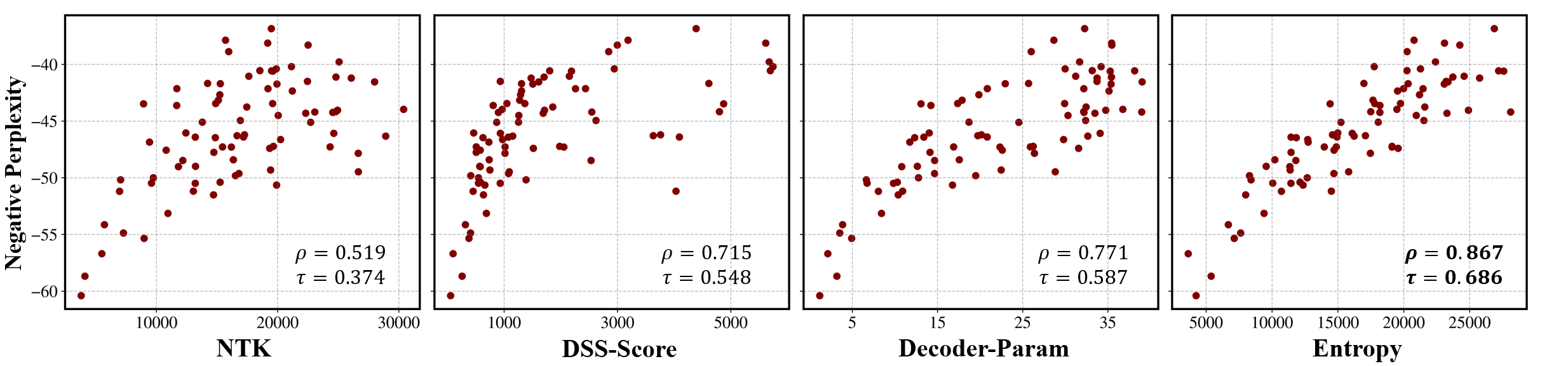}
         \vspace{-7mm}
        \caption{Correlation comparison of different training-free predictors, e.g., NTK~\cite{tenas}, DSS-Score~\cite{transformerfree}, and Decoder-Param~\cite{lts}, and transformer performance (negative perplexity, higher is better). $\rho$ is Spearman's Rank and $\tau$ is Kendall Tau. Larger values mean higher correlation. }
     \label{fig:corr-1}
\end{figure*}

\minisection{Evaluating Transformer without Training.} 
Recent studies~\cite{mae,maedet} have demonstrated that entropy, which captures the information capacity of neural network architecture, can be a reliable indicator for final model accuracy. 
In this part, we provide experimental results that empirically establish a strong correlation between our proposed entropy of transformers and their perplexity results on the One Billion Word (LM1B)~\cite{lm1b} dataset after training.
To this end, we uniformly sample 81 unique transformer architectures and each model is fully trained from \textit{scratch}. 
The performance is measured using perplexity and we calculate the correlation between each transformer model's entropy and perplexity score on the validation set.

Figure~\ref{fig:corr-1} illustrates the correlation between the sampled architectures' performance (negative perplexity) and their entropy. The results indicate strong correlations, as evidenced by Spearman's Rank Correlation ($\rho$) and Kendall Rank Correlation ($\tau$) scores exceeding 0.8 and 0.6, respectively. 
Furthermore, we compare our entropy to three commonly used low-cost evaluation proxies, namely Decoder Params in LTS~\cite{lts}, DSS-Score in TF-TAS~\cite{transformerfree}, and NTK in TE-NAS~\cite{tenas}.
As depicted in Figure~\ref{fig:corr-1}, our proposed entropy outperforms the other three accuracy predictors. 
Particularly, our method is more capable of identifying high-performance transformer architecture than previous methods.

Note that there are two principal distinctions between our entropy-driven approach and previous zero-shot NAS methods~\cite{lts,transformerfree,tenas}.
\underline{First}, zero-shot NAS methods are predominantly \textit{data-driven}. 
Our method, on the other hand, is mathematically driven with clear motivation from the perspective of information theory~\cite{mae}. \underline{Second}, zero-shot NAS methods usually require forward/backward passes over the architecture, where the model parameters and feature maps have to be stored in GPU memory. 
In contrast, our methodology is purely analytical and the expensive entropy calculation process is substituted by a table lookup procedure, therefore highly efficient. 
\textbf{Our method requires zero GPU memory in the design stage}. 
In summary, our method is a much better approach to designing efficient language models for edge devices than zero-shot NAS.

\begin{figure}[!t]
     \centering
         \includegraphics[width=\linewidth]{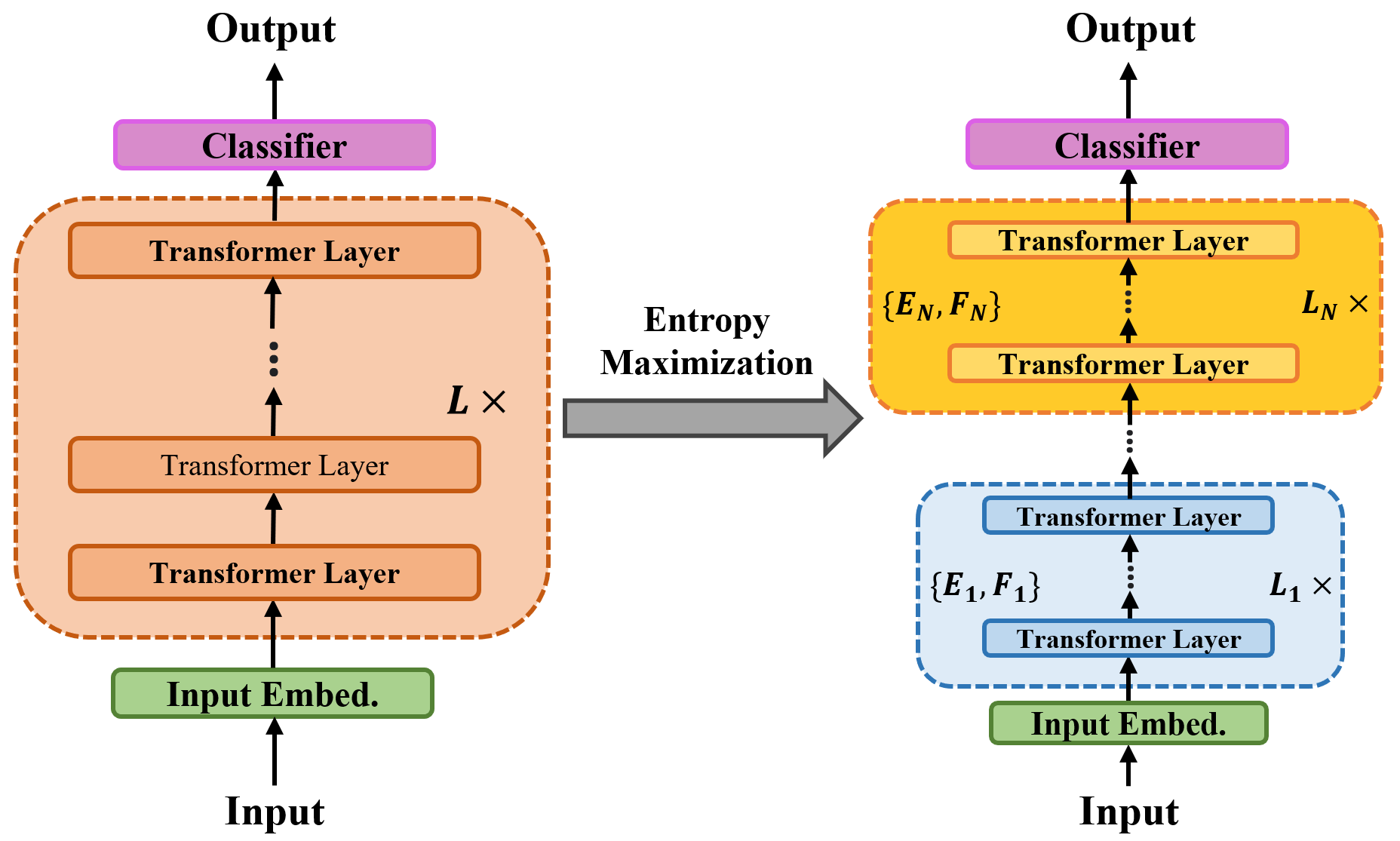}
         \vspace{-6mm}
        \caption{Our proposed adaptive block-wise transformer design. Left is the standard autoregressive transformer design, which consists of $L$ homogeneous layers, and right is the optimal architecture design after entropy maximization, where there are  $N$ number of transformer blocks and each transformer block has adaptive width (${E_i,R_i}$) and depth ($L_i$).}
     \label{fig:ss}
\end{figure}
\subsection{Designing Mobile Language Models}
\minisection{Search Space.} In the design of MeRino, we introduce an adaptive block-wise search space to construct the backbone architecture, as shown in Figure~\ref{fig:ss}.
Each transformer block consists of numerous transformer layers of the same number of attention heads, FFN dimensions, and embedding dimensions. 
Moreover, we incorporate parameter sharing technique~\cite{albert} within each transformer block. This means that all MHSA and FFN layers share the same weights, resulting in transformer models with reduced memory footprint.

Within each transformer block, in MHSA layers, we fix the head dimension to 64 and make the attention head number elastic so that each attention module can decide its necessary number of heads. 
We also set the \textit{Q-K-V} dimensions the same as embedding dimensions. 
To prevent information bottlenecks, we also ensure that as the network goes deeper, the embedding dimension of each transformer block should be non-decreasing.

\minisection{Search Process.} To design a transformer model $f(\cdot)$ with $N$ transformer blocks under a given computation budget $C$, we propose to optimize the parameters $\{E_j, F_j, L_j\}_{j=1,...,N}$ by solving a mathematical programming (MP) problem. $E_j$, $F_j$, and $L_j$ denote the embedding dimension, FFN dimension, and number of layers in the $j$-th transformer block, respectively. For simplicity, we use ComputeCost as a general term to represent computation constraints such as parameter size, FLOPs, and latency.

The objective of the MP problem is to maximize the total transformer entropy, representing the expressiveness and effectiveness of the model while considering constraints on the computational cost. The MP problem is formulated as follows:

\begin{align}
\label{eq:MP}
\begin{split}
    \max_{\{E_i, F_i, L_i\}}  &  \sum_{j=1}^N L_{j} [(1-\frac{\beta L_j}{\log E_j}) \widehat{H}_{\text{MHSA}} + (1-\frac{\beta L_j}{\log F_j}) \widehat{H}_{\text{FFN}}] \\
    \mathrm{s.t.} \quad & \mathrm{ComputeCost}[f(\cdot)] \leq \mathcal{C}, \quad  E_1 \leq \cdots \leq E_N
\end{split}
\end{align}

\renewcommand{\arraystretch}{1.25} 
\begin{table*}[!t]
\caption{Detailed zero-shot learning results for MeRino and publicly available pre-trained LLMs.}
\vspace{-3mm}
\resizebox{\linewidth}{!}{
\begin{tabular}{lcccccccccccccc}
\toprule
 & Params & HellaSwag & WinoGrande & ARC-E & ARC-C & OpenbookQA & BoolQ & WIC & CB & WSC & RTE & PubmedQA & LogiQA & Avg. \\ \midrule
Pythia-70M & 70\,M & 0.269 & \textbf{0.529} & 0.335 & 0.214 & 0.272 & 0.589 & 0.486 & 0.339 & 0.365 & 0.523 & 0.409 & 0.266 & 0.383 \\
Pythia-162M & 162\,M & 0.292 & 0.492 & 0.373 & 0.231 & 0.264 & 0.571 & 0.500 & 0.446 & 0.365 & \textbf{0.563} & 0.544 & 0.269 & \textbf{0.409}\\
Cerebras-111M & 111\,M & 0.267 & 0.490 & 0.336 & 0.207 & 0.256 & 0.621 & 0.500 & 0.411 & 0.365 & 0.549 & \textbf{0.552} & 0.266 & 0.402\\ 
GPT-2-124M & 124\,M & \textbf{0.300} & 0.516 & 0.382 & 0.230 & 0.272 & 0.554 & 0.492 & 0.410 & \textbf{0.433} & 0.531 & 0.430 & 0.245 & 0.400\\
OPT-125M & 125\,M & 0.267 & 0.503 & 0.386 & 0.233 & 0.226 & 0.554 & 0.498 & 0.357 & 0.365 & 0.444 & 0.372 & \textbf{0.286} & 0.373 \\
OPT-350M & 331\,M & 0.283 & 0.523 & \textbf{0.389} & 0.233 & \textbf{0.286} & 0.618 & 0.500 & \textbf{0.464} & 0.365 & 0.542 & 0.414 & 0.280 & 0.408\\ \midrule
MeRino-52M & 52\,M & 0.267 & 0.507 & 0.327 & 0.212 & 0.242 & 0.541 & \textbf{0.525} & \textbf{0.411} & 0.413 & 0.502 & 0.377 & 0.276 & 0.383\\
MeRino-61M & 61\,M & 0.273 & 0.510 & 0.336 & 0.209 & 0.248 & 0.610 & 0.502 & 0.375 & 0.365 & 0.534 & 0.484 & 0.255 & 0.392 \\
MeRino-64M & 64\,M & 0.274 & 0.528 & 0.341 & \textbf{0.234} & 0.267 & \textbf{0.621} & 0.505 & 0.393 & 0.375 & 0.545 & 0.540 & 0.278 & 0.408 \\ 
 \bottomrule
\end{tabular}
}
\label{tab:main}
\end{table*}

To solve this optimization problem, we employ an Evolutionary Algorithm~\cite{ea}. 
Note that Eq.~(\ref{eq:MP}) can be solved by any non-linear programming solver in principle.
We choose EA due to its simplicity. 
A detailed description of the EA algorithm and the mutation algorithm is given in the Appendix.

\begin{table}[!t]
\centering
\caption{Performance comparison on WikiText-2~\cite{wiki} and Penn TreeBank (PTB)~\cite{ptb} for language modeling tasks.}
\vspace{-2mm}
\resizebox{.92\linewidth}{!}
{\begin{tabular}{lcccc}
\toprule
 & FLOPs ($\downarrow$) & Latency ($\downarrow$) & \textbf{WikiText-2} & \textbf{PTB} \\ \midrule
Pythia-70M  & 100\,G & 95\,ms &40.95 & 60.28 \\ 
Pythia-162M  & 270\,G & 243\,ms &23.52 & 36.02 \\ 
Cerebras-111M  & 260\,G & 185\,ms &36.93 & 51.89 \\ 
GPT-2-124M  & 290\,G & 213\,ms &25.19 & 33.95 \\ 
OPT-125M  & 210\,G & 182\,ms &23.62 &  29.02 \\ 
OPT-350M  & 720\,G & 559\,ms &\textbf{18.51} & \textbf{23.08} \\ \midrule
MeRino-52M  & 60\,G & 48\,ms &39.05 & 52.18 \\ 
MeRino-61M  & 110\,G & 77\,ms &34.24 & 34.11 \\ 
MeRino-64M & 160\,G & 114\,ms &22.47 & 27.06 \\ 
\bottomrule
\end{tabular}
}
\label{tab:lm}
\end{table}

\section{Experiments}
In this section, we first describe detailed settings for search, training, and evaluation. 
Next, we report the main results of MeRino on various NLP tasks. 
Finally, we conduct ablation studies of each key component in our design methodology.
\subsection{Experimental Settings}
\minisection{Datasets.} For pre-training, we use the publicly available Pile dataset~\cite{pile}, which is pre-processed by removing duplication and tokenized using byte-level encoding.
For evaluation, we evaluate our models across fourteen different downstream NLP tasks, namely
WikiText-2~\cite{wiki}, Penn Treebank (PTB)~\cite{ptb},
HellaSwag~\citep{hellaswag}, WinoGrande~\citep{winogrande}, OpenBookQA~\citep{openbookqa}, ARC~\citep{arc}, PubmedQA~\citep{pubmedqa}, LogiQA~\citep{logiqa}, and SuperGLUE~\citep{superglue} benchmark BoolQ, CB, WIC, WSC and RTE.
We report the model performance using the perplexity and average accuracy for language modeling tasks, and zero-shot learning tasks, respectively.

\minisection{Implementation Details.} 
We follow the settings in~\cite{opt} and train each model from scratch for 600k steps with an effective batch size of 1024 and sequence length of 1024 on 8 NVIDIA H100 80\,GB GPUs. 
For the learning rate schedule, we follow~\cite{pythia} and adopt AdamW~\cite{adamw} optimizer, with a starting learning rate of 6e-4, warm-up steps of 1000, and linear learning rate decay. 
For evaluation, we adopt the codebase of lm-evaluation-harness~\cite{lm-eval} for a fair comparison.
We conduct latency experiments on NVIDIA Jetson Nano GPU 8GB and the inference latency is calculated with a batch size of 1 and sequence length of 128 averaged over 16 measurements.

\subsection{Main Results} 

\minisection{Comparison with SOTA LLMs.} As our scope is mobile-friendly language models, we mainly compare pre-trained LLMs under 1B parameters. 
At the time of paper submission, the weights and evaluation for MobileLLM~\cite{MobileLLM} are not released, thus we do not directly compare our method with it.

Table~\ref{tab:main} and \ref{tab:lm} report the detailed accuracy and latency results of our MeRino models and baseline models, such as GPT-2~\cite{gpt2}, OPT~\cite{opt}, Pythia~\cite{pythia} and Cerebras-GPT~\cite{cerebras}. Compared to the OPT family, MeRino achieves superior accuracy with much less parameter size and FLOPs. 
Specifically, MeRino-64M obtains similar average accuracy as OPT-350M but with 82\% and 78\% reduction in model size and computation respectively. 
Above all, MeRino achieves an average inference speedup of 4.6$\times$ against OPT family models, respectively. 
Our smallest model, MeRino-52M achieves similar performance as Pythia-70M but with 2.0$\times$ faster runtime.

\begin{figure}[!t]
     \centering
         \includegraphics[width=0.9\linewidth]{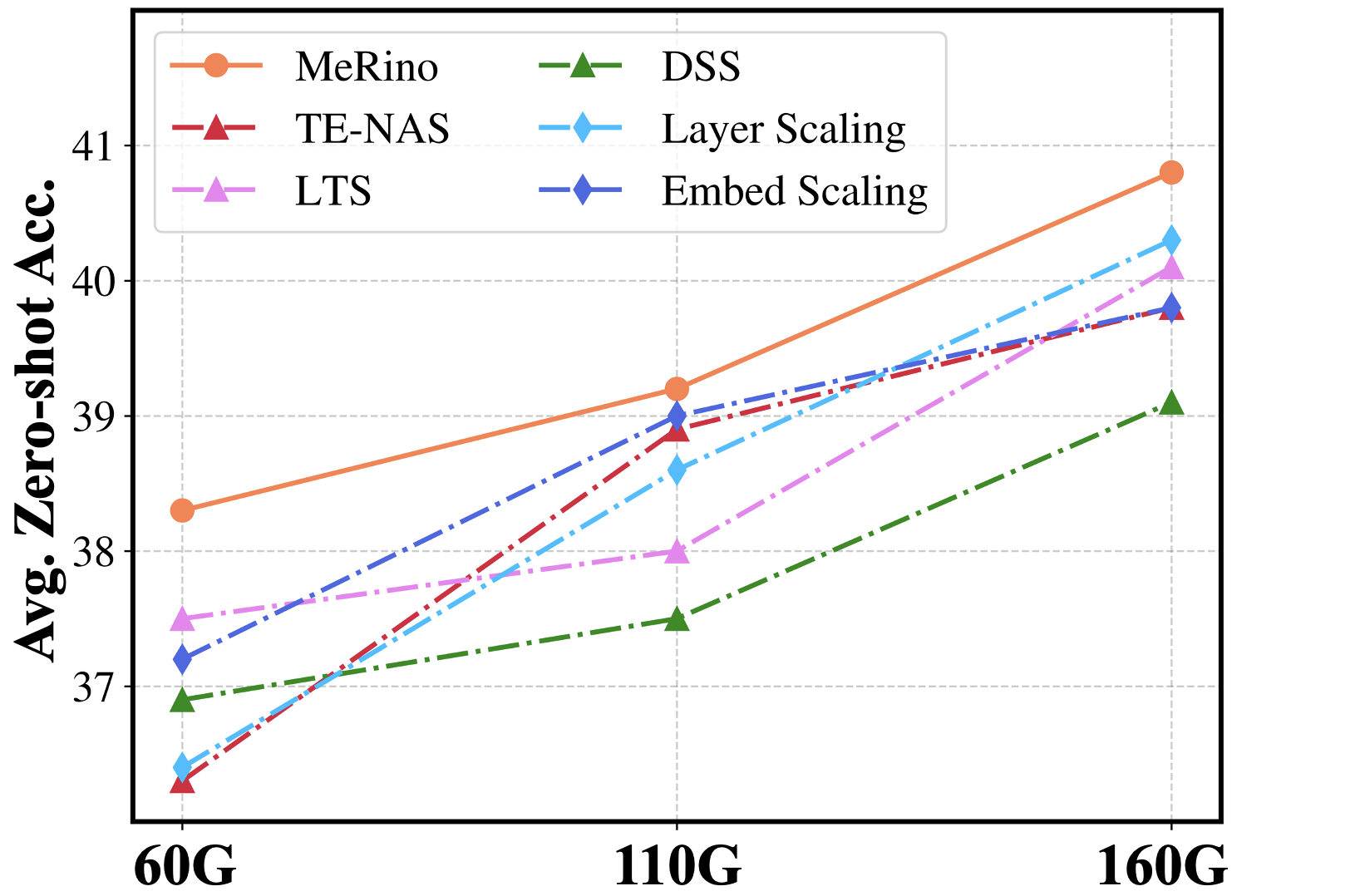}
         \vspace{-2mm}
        \caption{Performance comparison of MeRino, NAS-based methods, and naive scaling methods.
        }
     \label{fig:search}
\end{figure}
\minisection{Comparison with Zero-shot NAS.}
We compare our methods against three zero-shot NAS approaches, namely TE-NAS~\cite{tenas}, DSS-Score~\citep{transformerfree} and LTS~\citep{lts}.
We conduct searches using the same FLOPs constraints (60/110/160\,G), and report the downstream NLP performance of searched architectures. 
We also compare MeRino with naive scaling of the layer or embedding dimension for transformer-based LLMs.
In Figure~\ref{fig:search}, we can see that under the same computation constraint, our entropy-driven design can produce much more capable language models than both the naive scaling method and NAS-based approaches. 

We also compare the search efficiency of our method with previous zero-cost NAS.
Since the entropy-based score is based on pure analytical formulation, our method can directly run on target low-end hardware instead of GPU or TPU. 
In Table~\ref{tab:search}, we can see that it only takes about 0.05 hours to run on NVIDIA Jetson Nano, while TE-NAS consumes around 1.2 hours with a single NVIDIA GTX 1080Ti GPU. 
This further validates the high efficiency and scalability of our design methodology for diverse IoT devices.

\begin{figure}[!ht]
     \centering
         \includegraphics[width=0.8\linewidth]{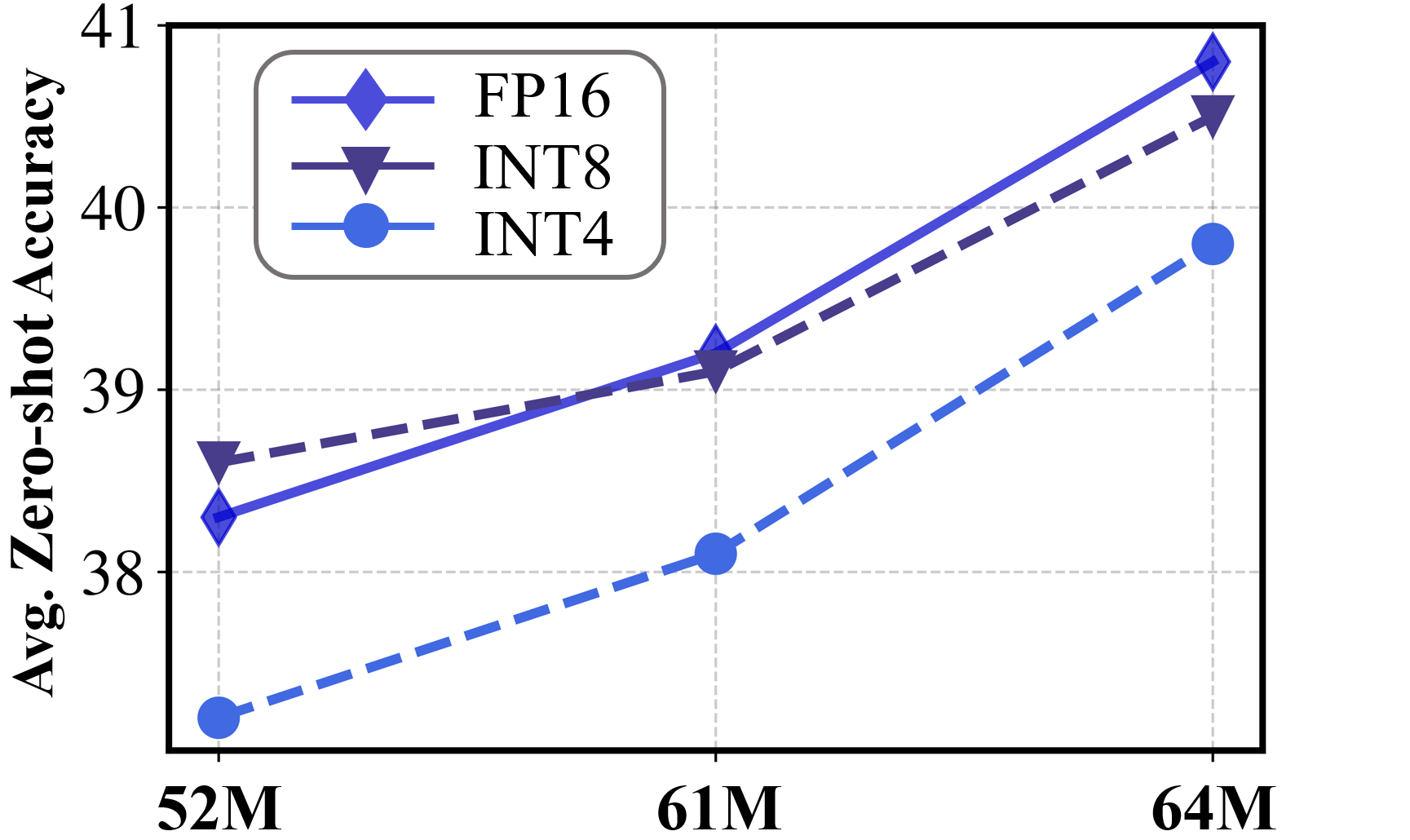}
         \vspace{-2mm}
        \caption{Comparison of our MeRino models in FP16, INT8 and INT4 precision.
        }
     \label{fig:quantization}
\end{figure}

\minisection{Combining Quantization.}
We further apply the popular weight quantization method, GPTQ~\cite{GPTQ} to our MeRino models. 
Figure~\ref{fig:quantization} shows that MeRino can maintain almost identical performance when applying INT8 quantization, and yields only an accuracy drop of 1\% in lower INT4 format.
\begin{table}[!t]
\centering
\caption{Searching cost comparison of TE-NAS and our method. *: NVIDIA GTX 1080Ti GPU; †: NVIDIA Jetson Nano.}
\vspace{-3mm}
\resizebox{.9\linewidth}{!}{
\begin{tabular}{rcccc}
\toprule
Method & \begin{tabular}[c]{@{}c@{}}Search\\ Device\end{tabular} & \begin{tabular}[c]{@{}c@{}}Search \\ Time (h)\end{tabular} & \begin{tabular}[c]{@{}c@{}}Energy\\ Costs (Wh)\end{tabular} & \begin{tabular}[c]{@{}c@{}}Average\\Acc.\end{tabular} \\ \midrule
TE-NAS & GPU* & 1.2 & 300 & 0.389 \\
\textbf{Ours} & CPU† & 0.05 & 0.75 & \textbf{0.408} \\ \bottomrule
\end{tabular}
}
\label{tab:search}
\end{table}

\subsection{Ablation Study}
\minisection{Impact of Parameter Sharing.} We further study the impact of the parameter sharing technique for MeRino, and present the results in Table~\ref{tab:ab-1}. We can see that sharing parameters within the same transformer block helps improve both parameter efficiency and downstream zero-shot average accuracy. For instance,  under the FLOPs budget of 160\,G, block-wise parameter sharing reduces parameter size by 36\% and improves downstream performances by 0.2\%. On average, with the parameter sharing technique, MeRino obtains 23\% reduction in parameter size and 0.5\% accuracy gain.

\minisection{Impact of $\gamma$ and $\alpha$.} As shown in Table~\ref{tab:ab-2}, effectiveness constraint $\gamma$ plays a key role in helping our entropy-driven framework design more capable and trainable models. 
When using effectiveness constraint $\gamma$, the final searched language model obtains +2.4\% average accuracy gain.
Similarly, we can see that using weighted entropy helps improve the average zero-shot accuracy by 0.8\%.
\begin{table}[!t]
\centering
\caption{Performance comparison of parameter sharing technique under three different FLOPs target.}
\label{tab:ab-1}
\vspace{-3mm}
\resizebox{.85\linewidth}{!}{
\begin{tabular}{cccc}
\toprule
Parameter Sharing & Params & FLOPs & \begin{tabular}[c]{@{}c@{}}Average\\Acc.\end{tabular} \\ \midrule
 & 59\,M & \multirow{2}{*}{60\,G} & 0.381 \\
\checkmark & 52\,M & & \textbf{0.383} \\ \midrule
 & 79\,M & \multirow{2}{*}{110\,G} & \textbf{0.395} \\ 
\checkmark & 61\,M &  & 0.392 \\ \midrule
 & 100\,M & \multirow{2}{*}{160\,G} & 0.403 \\
\checkmark & 64\,M & & \textbf{0.408}  \\  \bottomrule
\end{tabular}
}
\end{table}
\begin{table}[!t]
\centering
\caption{Accuracy performance comparison of effectiveness constraint and weighted entropy.}
\vspace{-3mm}
\label{tab:ab-2}
\resizebox{.9\columnwidth}{!}{
\begin{tabular}{cc|cc|c}
\toprule
\begin{tabular}[c]{@{}c@{}}Effectiveness\\ Constraint $\gamma$\end{tabular} & \begin{tabular}[c]{@{}c@{}}Weighted\\ Entropy $\alpha$\end{tabular} & \begin{tabular}[c]{@{}c@{}}Params\end{tabular} & \begin{tabular}[c]{@{}c@{}}FLOPs\end{tabular} & \begin{tabular}[c]{@{}c@{}}Average\\Acc.\end{tabular} \\ 
\midrule
&  & 62\,M & \multirow{3}{*}{110\,G} &  0.360 \\
 \checkmark &   & 59\,M &  & 0.384 \\
\checkmark & \checkmark  & 61\,M &  & \textbf{0.392} \\ 
 \bottomrule
\end{tabular}
}
\end{table}

\section{Conclusion}
In this paper, we propose a novel design framework aiming to generate efficient autoregressive language models for mobile devices under 100\,M parameters at nearly zero cost.
Leveraging the Maximum Entropy Principle, we formulate a constrained mathematical programming problem and optimize the network architecture by maximizing the entropy of transformer decoders under given computational budgets. 
Our designed model, MeRino can achieve comparable performance against both state-of-the-art pre-trained LLMs and NAS-designed models with significant improvement in model size reduction and inference runtime speedup on NVIDIA Jetson Nano. 

\section{Acknowledgment}
This work was sponsored in part by the U.S. National Science Foundation (NSF) under Grants 1907765 and 2400014.
The authors would like to thank the anonymous AAAI reviewers for their constructive feedback to improve this work.

\bibliography{aaai25}

\begin{thebibliography}{45}
\providecommand{\natexlab}[1]{#1}

\bibitem[{Anthropic(2023)}]{cluade}
Anthropic. 2023.
\newblock Introducting Claude.

\bibitem[{Apple(2024)}]{appleai}
Apple. 2024.
\newblock Apple Intelligence.

\bibitem[{Ashkboos et~al.(2024)Ashkboos, Croci, do~Nascimento, Hoefler, and Hensman}]{slicegpt}
Ashkboos, S.; Croci, M.~L.; do~Nascimento, M.~G.; Hoefler, T.; and Hensman, J. 2024.
\newblock SliceGPT: Compress Large Language Models by Deleting Rows and Columns.
\newblock \emph{ArXiv}, abs/2401.15024.

\bibitem[{Biderman et~al.(2023)Biderman, Schoelkopf, Anthony, Bradley, O'Brien, Hallahan, Khan, Purohit, Prashanth, Raff, Skowron, Sutawika, and van~der Wal}]{pythia}
Biderman, S.~R.; Schoelkopf, H.; Anthony, Q.~G.; Bradley, H.; O'Brien, K.; Hallahan, E.; Khan, M.~A.; Purohit, S.; Prashanth, U.~S.; Raff, E.; Skowron, A.; Sutawika, L.; and van~der Wal, O. 2023.
\newblock Pythia: A Suite for Analyzing Large Language Models Across Training and Scaling.

\bibitem[{Brown et~al.(2020)Brown, Mann, Ryder, Subbiah, Kaplan, Dhariwal, Neelakantan, Shyam, Sastry, Askell, Agarwal, Herbert-Voss, Krueger, Henighan, Child, Ramesh, Ziegler, Wu, Winter, Hesse, Chen, Sigler, Litwin, Gray, Chess, Clark, Berner, McCandlish, Radford, Sutskever, and Amodei}]{gpt3}
Brown, T.~B.; Mann, B.; Ryder, N.; Subbiah, M.; Kaplan, J.; Dhariwal, P.; Neelakantan, A.; Shyam, P.; Sastry, G.; Askell, A.; Agarwal, S.; Herbert-Voss, A.; Krueger, G.; Henighan, T.~J.; Child, R.; Ramesh, A.; Ziegler, D.~M.; Wu, J.; Winter, C.; Hesse, C.; Chen, M.; Sigler, E.; Litwin, M.; Gray, S.; Chess, B.; Clark, J.; Berner, C.; McCandlish, S.; Radford, A.; Sutskever, I.; and Amodei, D. 2020.
\newblock Language Models are Few-Shot Learners.
\newblock \emph{ArXiv}, abs/2005.14165.

\bibitem[{Chan et~al.(2021)Chan, Yu, You, Qi, Wright, and Ma}]{redunet}
Chan, K. H.~R.; Yu, Y.; You, C.; Qi, H.; Wright, J.; and Ma, Y. 2021.
\newblock ReduNet: A White-box Deep Network from the Principle of Maximizing Rate Reduction.
\newblock \emph{ArXiv}, abs/2105.10446.

\bibitem[{Chelba et~al.(2013)Chelba, Mikolov, Schuster, Ge, Brants, Koehn, and Robinson}]{lm1b}
Chelba, C.; Mikolov, T.; Schuster, M.; Ge, Q.; Brants, T.; Koehn, P.~T.; and Robinson, T. 2013.
\newblock One billion word benchmark for measuring progress in statistical language modeling.
\newblock In \emph{Interspeech}.

\bibitem[{Chen, Gong, and Wang(2021)}]{tenas}
Chen, W.; Gong, X.; and Wang, Z. 2021.
\newblock Neural Architecture Search on ImageNet in Four GPU Hours: A Theoretically Inspired Perspective.
\newblock \emph{ArXiv}, abs/2102.11535.

\bibitem[{Clark et~al.(2018)Clark, Cowhey, Etzioni, Khot, Sabharwal, Schoenick, and Tafjord}]{arc}
Clark, P.; Cowhey, I.; Etzioni, O.; Khot, T.; Sabharwal, A.; Schoenick, C.; and Tafjord, O. 2018.
\newblock Think you have Solved Question Answering? Try ARC, the AI2 Reasoning Challenge.
\newblock \emph{ArXiv}, abs/1803.05457.

\bibitem[{Dey et~al.(2023)Dey, Gosal, Chen, Khachane, Marshall, Pathria, Tom, and Hestness}]{cerebras}
Dey, N.; Gosal, G.~S.; Chen, Z.; Khachane, H.; Marshall, W.; Pathria, R.; Tom, M.; and Hestness, J. 2023.
\newblock Cerebras-GPT: Open Compute-Optimal Language Models Trained on the Cerebras Wafer-Scale Cluster.

\bibitem[{Frantar and Alistarh(2023)}]{sparsegpt}
Frantar, E.; and Alistarh, D. 2023.
\newblock SparseGPT: Massive Language Models Can Be Accurately Pruned in One-Shot.
\newblock \emph{ArXiv}, abs/2301.00774.

\bibitem[{Frantar et~al.(2023)Frantar, Ashkboos, Hoefler, and Alistarh}]{GPTQ}
Frantar, E.; Ashkboos, S.; Hoefler, T.; and Alistarh, D. 2023.
\newblock GPTQ: Accurate Post-Training Quantization for Generative Pre-trained Transformers.
\newblock \emph{International Conference on Learning Representations (ICLR)}.

\bibitem[{Gao et~al.(2020)Gao, Biderman, Black, Golding, Hoppe, Foster, Phang, He, Thite, Nabeshima, Presser, and Leahy}]{pile}
Gao, L.; Biderman, S.~R.; Black, S.; Golding, L.; Hoppe, T.; Foster, C.; Phang, J.; He, H.; Thite, A.; Nabeshima, N.; Presser, S.; and Leahy, C. 2020.
\newblock The Pile: An 800GB Dataset of Diverse Text for Language Modeling.
\newblock \emph{ArXiv}, abs/2101.00027.

\bibitem[{Gao et~al.(2021)Gao, Tow, Biderman, Black, DiPofi, Foster, Golding, Hsu, McDonell, Muennighoff, Phang, Reynolds, Tang, Thite, Wang, Wang, and Zou}]{lm-eval}
Gao, L.; Tow, J.; Biderman, S.; Black, S.; DiPofi, A.; Foster, C.; Golding, L.; Hsu, J.; McDonell, K.; Muennighoff, N.; Phang, J.; Reynolds, L.; Tang, E.; Thite, A.; Wang, B.; Wang, K.; and Zou, A. 2021.
\newblock A framework for few-shot language model evaluation.

\bibitem[{Hu et~al.(2024)Hu, Tu, Han, He, Cui, Long, Zheng, Fang, Huang, Zhao, Zhang, Thai, Zhang, Wang, Yao, Zhao, Zhou, Cai, Zhai, Ding, Jia, Zeng, Li, Liu, and Sun}]{minicpm}
Hu, S.; Tu, Y.; Han, X.; He, C.; Cui, G.; Long, X.; Zheng, Z.; Fang, Y.; Huang, Y.; Zhao, W.; Zhang, X.; Thai, Z.~L.; Zhang, K.; Wang, C.; Yao, Y.; Zhao, C.; Zhou, J.; Cai, J.; Zhai, Z.; Ding, N.; Jia, C.; Zeng, G.; Li, D.; Liu, Z.; and Sun, M. 2024.
\newblock MiniCPM: Unveiling the Potential of Small Language Models with Scalable Training Strategies.
\newblock \emph{ArXiv}, abs/2404.06395.

\bibitem[{Javaheripi et~al.(2022)Javaheripi, Shah, Mukherjee, Religa, Mendes, de~Rosa, Bubeck, Koushanfar, and Dey}]{lts}
Javaheripi, M.; Shah, S.; Mukherjee, S.; Religa, T.~L.; Mendes, C. C.~T.; de~Rosa, G.; Bubeck, S.; Koushanfar, F.; and Dey, D. 2022.
\newblock LiteTransformerSearch: Training-free On-device Search for Efficient Autoregressive Language Models.
\newblock \emph{ArXiv}, abs/2203.02094.

\bibitem[{Jaynes(1957)}]{mae}
Jaynes, E.~T. 1957.
\newblock Information Theory and Statistical Mechanics.
\newblock \emph{Physical Review}, 106: 620--630.

\bibitem[{Jin et~al.(2019)Jin, Dhingra, Liu, Cohen, and Lu}]{pubmedqa}
Jin, Q.; Dhingra, B.; Liu, Z.; Cohen, W.~W.; and Lu, X. 2019.
\newblock PubMedQA: A Dataset for Biomedical Research Question Answering.
\newblock In \emph{Conference on Empirical Methods in Natural Language Processing}.

\bibitem[{Lan et~al.(2019)Lan, Chen, Goodman, Gimpel, Sharma, and Soricut}]{albert}
Lan, Z.; Chen, M.; Goodman, S.; Gimpel, K.; Sharma, P.; and Soricut, R. 2019.
\newblock ALBERT: A Lite BERT for Self-supervised Learning of Language Representations.
\newblock \emph{ArXiv}, abs/1909.11942.

\bibitem[{Levine et~al.(2020)Levine, Wies, Sharir, Bata, and Shashua}]{dwi}
Levine, Y.; Wies, N.; Sharir, O.; Bata, H.; and Shashua, A. 2020.
\newblock The Depth-to-Width Interplay in Self-Attention.
\newblock \emph{arXiv: Learning}.

\bibitem[{Liu et~al.(2020)Liu, Cui, Liu, Huang, Wang, and Zhang}]{logiqa}
Liu, J.; Cui, L.; Liu, H.; Huang, D.; Wang, Y.; and Zhang, Y. 2020.
\newblock LogiQA: A Challenge Dataset for Machine Reading Comprehension with Logical Reasoning.
\newblock In \emph{International Joint Conference on Artificial Intelligence}.

\bibitem[{Liu et~al.(2024)Liu, Zhao, Iandola, Lai, Tian, Fedorov, Xiong, Chang, Shi, Krishnamoorthi, Lai, and Chandra}]{MobileLLM}
Liu, Z.; Zhao, C.; Iandola, F.~N.; Lai, C.; Tian, Y.; Fedorov, I.; Xiong, Y.; Chang, E.; Shi, Y.; Krishnamoorthi, R.; Lai, L.; and Chandra, V. 2024.
\newblock MobileLLM: Optimizing Sub-billion Parameter Language Models for On-Device Use Cases.
\newblock \emph{ArXiv}, abs/2402.14905.

\bibitem[{Loshchilov and Hutter(2017)}]{adamw}
Loshchilov, I.; and Hutter, F. 2017.
\newblock Decoupled Weight Decay Regularization.
\newblock In \emph{International Conference on Learning Representations}.

\bibitem[{Marcus, Santorini, and Marcinkiewicz(1993)}]{ptb}
Marcus, M.~P.; Santorini, B.; and Marcinkiewicz, M.~A. 1993.
\newblock Building a Large Annotated Corpus of English: The Penn Treebank.
\newblock \emph{Comput. Linguistics}, 19: 313--330.

\bibitem[{Merity et~al.(2016)Merity, Xiong, Bradbury, and Socher}]{wiki}
Merity, S.; Xiong, C.; Bradbury, J.; and Socher, R. 2016.
\newblock Pointer Sentinel Mixture Models.
\newblock \emph{ArXiv}, abs/1609.07843.

\bibitem[{Microsoft(2024)}]{copilot}
Microsoft. 2024.
\newblock Introducing Copilot+ PCs.

\bibitem[{Mihaylov et~al.(2018)Mihaylov, Clark, Khot, and Sabharwal}]{openbookqa}
Mihaylov, T.; Clark, P.; Khot, T.; and Sabharwal, A. 2018.
\newblock Can a Suit of Armor Conduct Electricity? A New Dataset for Open Book Question Answering.
\newblock In \emph{Conference on Empirical Methods in Natural Language Processing}.

\bibitem[{OpenAI(2022)}]{chatgpt}
OpenAI. 2022.
\newblock Introducting ChatGPT.

\bibitem[{Radford et~al.(2019)Radford, Wu, Child, Luan, Amodei, and Sutskever}]{gpt2}
Radford, A.; Wu, J.; Child, R.; Luan, D.; Amodei, D.; and Sutskever, I. 2019.
\newblock Language Models are Unsupervised Multitask Learners.

\bibitem[{Reeves(2007)}]{ea}
Reeves, C.~R. 2007.
\newblock Evolutionary computation: a unified approach.
\newblock \emph{Genetic Programming and Evolvable Machines}, 8: 293--295.

\bibitem[{Roberts, Yaida, and Hanin(2021)}]{dlt}
Roberts, D.~A.; Yaida, S.; and Hanin, B. 2021.
\newblock The Principles of Deep Learning Theory.
\newblock \emph{ArXiv}, abs/2106.10165.

\bibitem[{Sakaguchi et~al.(2019)Sakaguchi, Bras, Bhagavatula, and Choi}]{winogrande}
Sakaguchi, K.; Bras, R.~L.; Bhagavatula, C.; and Choi, Y. 2019.
\newblock WINOGRANDE: An Adversarial Winograd Schema Challenge at Scale.
\newblock \emph{Commun. ACM}, 64: 99--106.

\bibitem[{Saxe et~al.(2018)Saxe, Bansal, Dapello, Advani, Kolchinsky, Tracey, and Cox}]{info1}
Saxe, A.~M.; Bansal, Y.; Dapello, J.; Advani, M.~S.; Kolchinsky, A.; Tracey, B.~D.; and Cox, D.~D. 2018.
\newblock On the information bottleneck theory of deep learning.
\newblock \emph{Journal of Statistical Mechanics: Theory and Experiment}, 2019.

\bibitem[{Soham(2024)}]{query-energy}
Soham. 2024.
\newblock The Cost of Inference: Running the Models.

\bibitem[{Sun et~al.(2021)Sun, Lin, Sun, Tan, Li, and Jin}]{maedet}
Sun, Z.; Lin, M.; Sun, X.; Tan, Z.; Li, H.; and Jin, R. 2021.
\newblock MAE-DET: Revisiting Maximum Entropy Principle in Zero-Shot NAS for Efficient Object Detection.
\newblock In \emph{International Conference on Machine Learning}.

\bibitem[{Touvron et~al.(2023)Touvron, Lavril, Izacard, Martinet, Lachaux, Lacroix, Rozi{\`e}re, Goyal, Hambro, Azhar, Rodriguez, Joulin, Grave, and Lample}]{llama}
Touvron, H.; Lavril, T.; Izacard, G.; Martinet, X.; Lachaux, M.-A.; Lacroix, T.; Rozi{\`e}re, B.; Goyal, N.; Hambro, E.; Azhar, F.; Rodriguez, A.; Joulin, A.; Grave, E.; and Lample, G. 2023.
\newblock LLaMA: Open and Efficient Foundation Language Models.
\newblock \emph{ArXiv}, abs/2302.13971.

\bibitem[{Vaswani et~al.(2017)Vaswani, Shazeer, Parmar, Uszkoreit, Jones, Gomez, Kaiser, and Polosukhin}]{attention}
Vaswani, A.; Shazeer, N.~M.; Parmar, N.; Uszkoreit, J.; Jones, L.; Gomez, A.~N.; Kaiser, L.; and Polosukhin, I. 2017.
\newblock Attention is All you Need.
\newblock \emph{ArXiv}, abs/1706.03762.

\bibitem[{Wang et~al.(2019)Wang, Pruksachatkun, Nangia, Singh, Michael, Hill, Levy, and Bowman}]{superglue}
Wang, A.; Pruksachatkun, Y.; Nangia, N.; Singh, A.; Michael, J.; Hill, F.; Levy, O.; and Bowman, S.~R. 2019.
\newblock SuperGLUE: A Stickier Benchmark for General-Purpose Language Understanding Systems.
\newblock \emph{ArXiv}, abs/1905.00537.

\bibitem[{Wu et~al.(2022)Wu, Raghavendra, Gupta, Acun, Ardalani, Maeng, Chang, Aga, Huang, Bai et~al.}]{sustainableai}
Wu, C.-J.; Raghavendra, R.; Gupta, U.; Acun, B.; Ardalani, N.; Maeng, K.; Chang, G.; Aga, F.; Huang, J.; Bai, C.; et~al. 2022.
\newblock Sustainable ai: Environmental implications, challenges and opportunities.
\newblock \emph{Proceedings of Machine Learning and Systems}, 4: 795--813.

\bibitem[{Xu et~al.(2021)Xu, Tan, Luo, Song, Li, Qin, and Liu}]{nasbert}
Xu, J.; Tan, X.; Luo, R.; Song, K.; Li, J.; Qin, T.; and Liu, T.-Y. 2021.
\newblock NAS-BERT: Task-Agnostic and Adaptive-Size BERT Compression with Neural Architecture Search.
\newblock \emph{Proceedings of the 27th ACM SIGKDD Conference on Knowledge Discovery \& Data Mining}.

\bibitem[{Zellers et~al.(2019)Zellers, Holtzman, Bisk, Farhadi, and Choi}]{hellaswag}
Zellers, R.; Holtzman, A.; Bisk, Y.; Farhadi, A.; and Choi, Y. 2019.
\newblock HellaSwag: Can a Machine Really Finish Your Sentence?
\newblock In \emph{Annual Meeting of the Association for Computational Linguistics}.

\bibitem[{Zhang et~al.(2022)Zhang, Roller, Goyal, Artetxe, Chen, Chen, Dewan, Diab, Li, Lin, Mihaylov, Ott, Shleifer, Shuster, Simig, Koura, Sridhar, Wang, and Zettlemoyer}]{opt}
Zhang, S.; Roller, S.; Goyal, N.; Artetxe, M.; Chen, M.; Chen, S.; Dewan, C.; Diab, M.; Li, X.; Lin, X.~V.; Mihaylov, T.; Ott, M.; Shleifer, S.; Shuster, K.; Simig, D.; Koura, P.~S.; Sridhar, A.; Wang, T.; and Zettlemoyer, L. 2022.
\newblock OPT: Open Pre-trained Transformer Language Models.
\newblock \emph{ArXiv}, abs/2205.01068.

\bibitem[{Zhang et~al.(2023)Zhang, Sheng, Zhou, Chen, Zheng, Cai, Song, Tian, R{\'e}, Barrett, Wang, and Chen}]{h2o}
Zhang, Z.~A.; Sheng, Y.; Zhou, T.; Chen, T.; Zheng, L.; Cai, R.; Song, Z.; Tian, Y.; R{\'e}, C.; Barrett, C.~W.; Wang, Z.; and Chen, B. 2023.
\newblock H2O: Heavy-Hitter Oracle for Efficient Generative Inference of Large Language Models.
\newblock \emph{ArXiv}, abs/2306.14048.

\bibitem[{Zhao, Wu, and Wang(2024)}]{alisa}
Zhao, Y.; Wu, D.; and Wang, J. 2024.
\newblock ALISA: Accelerating Large Language Model Inference via Sparsity-Aware KV Caching.
\newblock \emph{2024 ACM/IEEE 51st Annual International Symposium on Computer Architecture (ISCA)}, 1005--1017.

\bibitem[{Zhou et~al.(2022)Zhou, Sheng, Zheng, Li, Sun, Tian, Chen, and Ji}]{transformerfree}
Zhou, Q.; Sheng, K.; Zheng, X.; Li, K.; Sun, X.; Tian, Y.; Chen, J.; and Ji, R. 2022.
\newblock Training-free Transformer Architecture Search.
\newblock \emph{2022 IEEE/CVF Conference on Computer Vision and Pattern Recognition (CVPR)}, 10884--10893.

\end{thebibliography}
\newpage
\section{Appendix}
\subsection{Evolutionary Algorithm}
We give a detailed description of the evolutionary algorithm (EA) and mutation algorithm in Algorithm~\ref{alg:ea} and Algorithm~\ref{alg:mutate}, respectively.

In designing MeRino models, we set the number of iterations $T$ to 100000, with a population size $M$ of 512 and the parent size $K$ of 64. We conduct searches for three different FLOP targets (60/110/160\,G). We limit the number of transformer blocks to $N=4$ and set $\beta=1/16$. 

\begin{algorithm}[!ht]
\caption{Evolutionary Algorithm}
\label{alg:ea}
\begin{algorithmic}
\Require Search space ${D}$, number of iterations $T$, computation budget ${C}$, population size $M$, parent size $K$
\Ensure Optimal architecture ${A}^*$
\State Initialize population ${P}$
\While{$i \leq T$}
\While{$len({P}) < M$}
\State Random select: ${A}_i \in {P}$ as parent.
\State Mutate: $\hat{{A}_i} = \textbf{MUTATE}({A}_i, {D})$
\If {$\text{ComputeCost}(\hat{{A}_i}) \leq {C}$}
\State Calculate entropy: $Z = H(\hat{{A}_i})$
\State Add $\hat{{A}_i}$ to ${P}$
\Else
\State Do nothing
\EndIf
\EndWhile
\State Remove $(M-K)$ networks with smallest entropy scores
\EndWhile
\State Return ${A}^*$, the architecture with the highest entropy in $P$
\end{algorithmic}
\end{algorithm}
\vspace{-6mm}
\begin{algorithm}[!ht]
\caption{MUTATE}
\label{alg:mutate}
\begin{algorithmic}
\Require Search space ${D}$, architecture ${A}_i$.
\Ensure Mutated architecture $\hat{{A}_i}$
\State Randomly select a block in ${A}_i$
\State Randomly alternate block depth, embedding dimension, and FFN ratio within a certain range
\State Return the mutated architecture $\hat{{A}_i}$
\end{algorithmic}
\end{algorithm}

\subsection{Search Space}
Table~\ref{tab:search} presents details of our search space for MeRino. In addition, we set the embedding projection dimension as 768 and the maximum position embedding dimension as 2048. Our search space encapsulates over 216k different autoregressive transformer architectures.
\begin{table}[!ht]
\centering
\caption{Search space configurations for MeRino.}
\vspace{-3mm}
\resizebox{.9\linewidth}{!}{
\begin{tabular}{cc}
\toprule
Embedding Dimension - $E_i$ & $\{64, ..., 1024|64\}$ \\
FFN Dimension - $F_i$ & $\{128, ..., 4096|128\}$ \\
Number of Layers Per Block - $L_i$ & $\{1,2,3,4\}$ \\ \bottomrule
\end{tabular}
}
\label{tab:search}
\end{table}

\subsection{Detail Structure of MeRino}
The searched network structures of MeRino are listed in Tables \ref{tab:structure}. 
We use four blocks for our MeRino design. $E_i$ denotes the embedding dimension for each transformer block, $R_i$ denotes the FFN ratio, and $L_i$ denotes the number of layers (depth) of each transformer block.
\begin{table}[ht]
\centering
\caption{Structure Configuration of MeRino.}
\vspace{-3mm}
\label{tab:structure}
\resizebox{\linewidth}{!}{
\begin{tabular}{cccc}
\toprule
Model & $E_i$ & $R_i$ & $L_i$ \\ \midrule
MeRino-52M & [512, 512, 640, 896] & [1, 1, 1, 1] & [2, 3, 2, 1] \\
MeRino-51M & [640, 768, 896, 1024] & [1, 1.5, 1, 1] & [2, 2, 2, 2]\\
MeRino-64M & [640, 896, 1024, 1024] & [1.5, 1.5, 1, 1] & [3, 3, 2, 3] \\ 
 \bottomrule
\end{tabular}
}
\end{table}

\end{document}